%% file: 00_main.tex
\documentclass[letterpaper, conference]{IEEEtran}
\IEEEoverridecommandlockouts
\usepackage{cite}
\usepackage{amsmath,amssymb,amsfonts}
\usepackage{graphicx}
\usepackage{textcomp}
\usepackage{xcolor}
\usepackage{pgfplotstable}
\usepackage{booktabs}

\usepackage{float}

\input{macros}

\def\BibTeX{{\rm B\kern-.05em{\sc i\kern-.025em b}\kern-.08em
    T\kern-.1667em\lower.7ex\hbox{E}\kern-.125emX}}

\begin{document}

\title{F$in$A: Fairness of Adverse Effects in Decision-Making of Human-Cyber-Physical-System
}

\author{\IEEEauthorblockN{Tianyu Zhao}
\IEEEauthorblockA{\textit{University of California, Irvine}\\
Irvine, CA \\
tzhao15@uci.edu}

\and
\IEEEauthorblockN{Salma Elmalaki}
\IEEEauthorblockA{\textit{University of California, Irvine}\\
Irvine, CA \\
salma.elmalaki@uci.edu}
}

% \and
% \and
% \and
% \and

\maketitle

%%% Remove this in the submission - enforcing page numbers
% \thispagestyle{plain}
% \pagestyle{plain}

\begin{abstract}

Ensuring fairness in decision-making systems within Human-Cyber-Physical-Systems (HCPS) is a pressing concern, particularly when diverse individuals, each with varying behaviors and expectations, coexist within the same application space, influenced by a shared set of control actions in the system. The long-term adverse effects of these actions further pose the challenge, as historical experiences and interactions shape individual perceptions of fairness. This paper addresses the challenge of fairness from an equity perspective of adverse effects, taking into account the dynamic nature of human behavior and evolving preferences while recognizing the lasting impact of adverse effects. We formally introduce the concept of Fairness-in-Adverse-Effects (F\emph{in}A) within the HCPS context. We put forth a comprehensive set of five formulations for F\emph{in}A, encompassing both the instantaneous and long-term aspects of adverse effects. 
To empirically validate the effectiveness of our F\emph{in}A approach, we conducted an evaluation within the domain of smart homes, a pertinent HCPS application. The outcomes of our evaluation demonstrate that the adoption of F\emph{in}A significantly enhances the overall perception of fairness among individuals, yielding an average improvement of $\textbf{66.7\%}$ when compared to the state-of-the-art method.

\end{abstract}


\input{01_introduction}

\input{02_relatedwork}
\input{03_approach}

\input{04_application_house}

\input{05_conclusion}

\section*{Acknowledgment}
This research was partially supported by NSF award \# CNS-2105084.

\clearpage
\bibliographystyle{IEEEtran}
\bibliography{citations}

\newpage
\appendices

\end{document}

%% file: macros.tex
\usepackage{xspace}
\usepackage{enumitem}
\usepackage{tabularx}
\usepackage{soul}
\usepackage{dsfont}
\usepackage{amsthm}
\usepackage{algorithm}
\usepackage[noend]{algpseudocode}
\usepackage{amsmath}

\usepackage[table]{colortbl}

\usepackage[font=small, skip=0.5ex]{caption}
\usepackage{caption}

\newcommand{\rom}[1]{\uppercase\expandafter{\romannumeral #1\relax}}

\usepackage{multirow}

%% file: 01_introduction.tex
\section{Introduction}

% The seamless interaction between humans and smart technologies represents a significant opportunity, but it also presents a daunting set of challenges. Understanding the intricate interactions between CPS and humans is essential for shaping the ultimate societal outcomes of future CPS technologies~\cite{george2023roadmap, Annaswamy2023}. 
The ubiquitous integration of smart technologies into our daily lives offers unprecedented opportunities but also presents a lot of challenges. A key challenge lies in understanding the interplay between humans and Cyber-Physical Systems (CPS) to shape the societal consequences of future CPS technologies.
As we move towards a future defined by the coexistence of humans and smart technologies, understanding their interactions is essential, as suggested by recent studies~\cite{george2023roadmap, Annaswamy2023}. Within the realm of Human-Cyber-Physical Systems (HCPS), one central challenge emerges when CPS decisions can affect individuals with diverse preferences and perceptions within the same environment. This scenario is common in systems like smart buildings, smart cities, smart traffic management, and smart crowd control, where fairness, privacy, equity, and personalization issues intersect, sparking new societal tensions~\cite{lee2017human,wang2020factors,stangor2014conflict}.

Existing research in sociology, particularly Social Exchange Theory~\cite{homans1974social,cook1987social} and Equity Theory~\cite{adams1963towards} has substantiated a direct link between ``equity'' and prosocial behavior. The higher an individual perceives a system as equitable, the greater the likelihood of that individual engaging in prosocial behavior~\cite{thibaut1959social,redmond2015social, adams1963towards}. This, in turn, significantly impacts the overall performance of the system. Specifically, the greater the number of individuals who engage in prosocial behavior, the higher the likelihood of compliance and acceptance of the system's decisions, ultimately leading to an enhancement in overall system performance~\cite{cialdini1987influence,cialdini2016pre,huijts2012psychological}. Given that CPS decision-making often aims to optimize system performance while adhering to operational constraints, it is essential to develop formal metrics and objective functions that enhance the human perception of these decisions, ultimately fostering more prosocial behavior and improving overall system performance.

In this paper, we are interested in formalizing some of the equity objectives in decision-making HCPS. We will start by exploring a notion of fairness from the equity perspective. In particular, we will focus on what we term \textbf{Fairness-in-Adverse-Effects (F\emph{in}A)}.  Decision-making agents in HCPS employ a range of control actions. However, these control actions can have different adverse effects on a diverse population, each with its own preferences.  
Hence, the HCPS needs to adapt its decision-making to continuously match different populations across time and ensure that it meets the preferences of as many populations as possible. Our motivation in examining the adverse effects or the harmful impact of decision-making in HCPS stems from the psychology concept \textbf{``loss aversion}--Losses loom larger than gains'' which implies that losses can be twice as powerful, psychologically, as gains~\cite{kahneman2013prospect}. 
This concept underscores the significance of minimizing adverse effects to promote fairness within HCPS.

%% file: 02_relatedwork.tex
\section{Related Work}
% \paragraph{Fairness in decision making} ....
% \paragraph{Fairness in Sequential decision making} ....

HCPS systems are centered on the challenge of designing adaptable, real-time decision-making processes that take into account the social context, including considerations of fairness, social welfare, ethical concerns, and societal norms~\cite{sztipanovits2019science, khargonekar2020framework}. A substantial body of work in the field of game theory explores various facets of fairness, often framed as incentive markets among competing entities or communities striving to achieve fairness~\cite{ratliff2020adaptive, ratliff2018perspective}. In the domain of machine learning, interventions to enhance fairness have been introduced, aiming to ensure that models' decisions are devoid of discrimination~\cite{friedler2019comparative, hashimoto2018fairness, chouldechova2018frontiers, kannan2018smoothed, goel2018non}. In the context of decision-making systems, where agents express favoritism for one action over another, questions surrounding fairness become even more significant, especially within multi-agent systems~\cite{jabbari2017fairness, joseph2016fairness, yu2019deep, gillen2018online, siddique2020learning, shin2017exploring, jiang2019learning, hughes2018inequity}. However, imposing fairness constraints as static, one-time decisions akin to conventional supervised learning methods while neglecting dynamic feedback and long-term consequences, particularly in sequential decision-making systems, can inadvertently lead to disparities that affect specific sub-populations~\cite{creager2020causal, liu2018delayed}.%, d2020fairness}.

Recent research has also shed light on the long-term ramifications of Reinforcement Learning (RL), revealing that addressing control decisions' immediate effects in single steps does not ensure fairness in subsequent decision actions~\cite{kannan2019downstream, milli2019social}. Nevertheless, a significant portion of this research has predominantly focused on fairness through the lens of equality, with an emphasis on eliminating favoritism or bias within the system, and relatively less attention has been given to the concept of fairness from an equity perspective, particularly in the context of sequential decision-making~\cite{mehrabi2021survey}. Notably, ensuring fairness in sequential decision-making systems becomes increasingly complex as policies deemed fair at one point may inadvertently become discriminatory over time due to shifts in human preferences influenced by the inter- and intra-human variation~\cite{elmalaki2021fair}. 

The concept of ``group fairness'' has been introduced in the literature to tackle fairness concerns arising when the same adaptive model impacts multiple individuals. One notion is ``Equalized Odds'' which concentrates on achieving a level of uniform prediction accuracy across various groups, primarily within the context of binary classification tasks. The central objective is to ensure that a predictive model exhibits comparable true positive rates (sensitivity) and true negative rates (specificity) across diverse groups~\cite{hardt2016equality}. A second notion is ``Equal opportunity,'', which aims to guarantee that a predictive model affords an equal likelihood of beneficial outcomes for all groups. It places a specific requirement on the true positive rate, mandating that it should be approximately equivalent for each group~\cite{hardt2016equality}.

Prior research in CPS has explored fairness in various ways. For instance, fairness-aware resource allocation and scheduling algorithms have been developed for CPS, addressing issues like energy consumption and real-time constraints while considering equitable distribution among system components~\cite{iosup2017self}. %~\cite{abdallah2017fair, michoud2015fair}. 
The concept of fairness in communication protocols for CPS has been established through strategies to ensure fair access to network resources for different devices and applications~\cite{huaizhou2013fairness}. %~\cite{verde2016fairness, lin2016fairness}. 
Furthermore, fairness challenges in decentralized CPS environments have been examined, focusing on ensuring fair decision-making in multi-agent systems~\cite{ho2020decentralized}. %~\cite{corre2016decentralized, fagiolini2017fair}.

However, it's worth noting that much of the existing CPS literature concentrates on system-level performance and efficiency, often at the cost of individual-level fairness considerations. In contrast, this paper aims to delve deeper into the aspects of fairness within HCPS, addressing the interplay between human preferences, the temporal dimension of adverse effects, and perceptions of fairness. This approach allows for a more comprehensive understanding of fairness in the context of HCPS decision-making, particularly in systems where individuals' preferences and perceptions can evolve over time.

\subsection{Paper contribution} 
This paper's contributions can be summarized as follows:
\begin{itemize}[itemsep=1pt]
\item \textbf{Fairness-in-Adverse-Effect (F\emph{in}A}): In this paper, we introduce a novel concept known as ``Fairness-in-Adverse-Effect (F\emph{in}A).'' F\emph{in}A takes a fresh perspective by considering equity in adverse effects within Human-Cyber-Physical Systems (HCPS) decision-making. We address scenarios where adaptive decisions made within HCPS can impact multiple individuals with diverse preferences. By formalizing F\emph{in}A, we provide a means to ensure that the adverse effects of decision-making are distributed fairly among the system's users

%\item \textbf{Long-term effects:} We formalize F\emph{in}A within CPS decision-making to capture the intricate interplay between human preference, the temporal dimension of adverse effects, and perceptions of fairness.

\item \textbf{Long-term effects:} Our work extends beyond the immediate effect of CPS decision-making. We delve into the temporal dimension of adverse effects, recognizing that the impact of these decisions can have lasting consequences. The interplay between human preferences, historical data, and the evolving perception of fairness adds complexity to the notion of fairness. We introduce five different approaches to formalize F\emph{in}A within CPS decision-making to account for the relationship between the instantaneous  adverse effects, long-term adverse effects, and the overall perception of fairness.

% \item \textbf{Generalization to different HCPS setups:} We provide a general formalization of F\emph{in}A and conduct comprehensive evaluations in the domain of smart buildings, demonstrating differences and trade-offs between various F\emph{in}A notions.%~\cite{elmalaki2021fair}.% and effectiveness in promoting fairness.

\item \textbf{Generalization to different HCPS setups:} We acknowledge that the nature of adverse effects and fairness considerations can vary across different domains. Therefore, we offer a general formalization of F\emph{in}A that can be applied to various HCPS scenarios. Additionally, we conduct thorough evaluations in the domain of smart home to illustrate the trade-offs between various interpretations and implementations of F\emph{in}A. This demonstrates the flexibility and effectiveness of our approach across diverse HCPS settings.
\end{itemize}

The rest of the paper is organized as follows: We first introduce the notion of Fairness-in-Adverse-Effects (F\emph{in}A) within Human-Cyber-Physical Systems (HCPS) in Section~\ref{sec:approach}. We formally define F$in$A using five different approaches in considering the instantaneous and the long-term adverse effects while considering the human perception of fairness in Sections~\ref{sec:minfina},~\ref{sec:minl}, ~\ref{sec:minfina+ul},~\ref{sec:miny} and~\ref{sec:miny+b}. Afterward, we numerically evaluate these approaches using a smart home HCPS application while comparing with the state-of-the-art in Section~\ref{sec:evaluation}.

%% file: 03_approach.tex
\section{Approach}\label{sec:approach}
\begin{figure}[!t]
\centering
\includegraphics[width=\columnwidth]{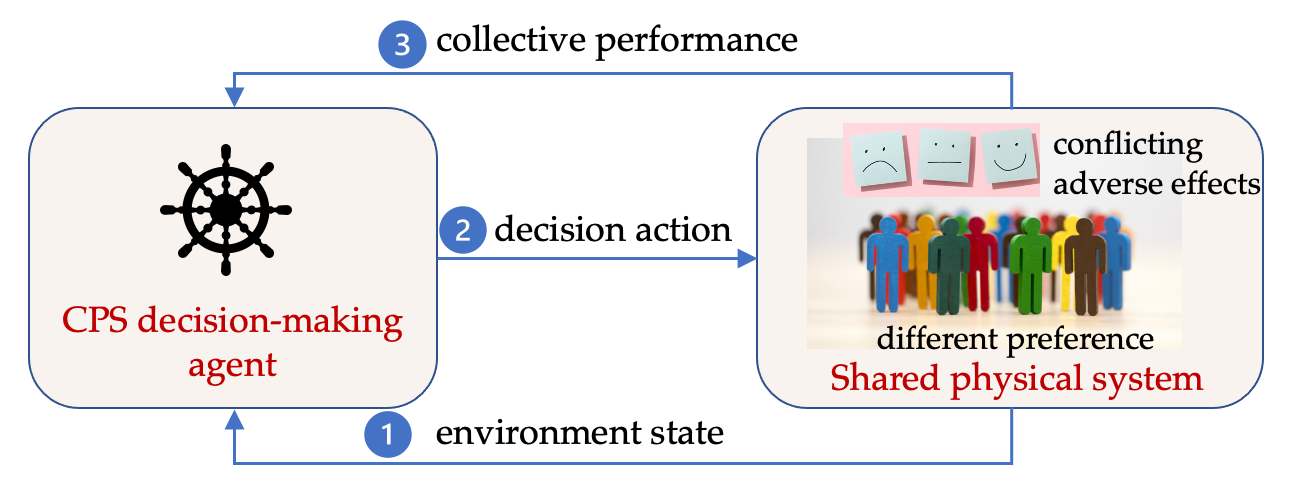}
\caption{Human-Cyber-Physical-System with multiple individuals sharing the same environments with different preferences. The decision-making agent control action can cause different adverse effects on those individuals.}\label{fig:hcps}
\end{figure}

We consider a CPS depicted in Figure~\ref{fig:hcps}, which serves multiple individuals sharing the same CPS environment, each with different preferences. The control action generated by the decision-making agent in CPS can cause different adverse effects on those individuals.

To achieve Fairness-in-Adverse-Effects (F\emph{in}A) within Human-Cyber-Physical Systems (HCPS), we propose five distinct approaches to guide CPS decision-making when selecting control actions affecting multiple humans sharing the same environment. These approaches are rooted in the recognition that individuals who exhibit pro-social attitudes might be willing to tolerate certain discrepancies between their preferences and CPS actions for a limited duration~\cite{thibaut1959social}. However, it is important to acknowledge that this willingness to forgive may not be indefinite, as the magnitude and persistence of discrepancies play a pivotal role in shaping individuals' perception of fairness~\cite{thibaut1959social}. Moreover, the extent of an individual's willingness to forgive is contingent upon the CPS's responsiveness in addressing and rectifying such discrepancies~\cite{cook1987social}.

Our first approach formalizes F\emph{in}A by examining the concept of instantaneous adverse effects, focusing on the immediate impact of CPS control actions (Section~\ref{sec:minfina}). In this context, we recognize that individuals may exhibit a degree of patience when faced with minor or unintentional discrepancies between their preferences and CPS actions. This approach is predicated on the assumption that in scenarios of minimal and short-term discrepancies, individuals may still perceive the CPS as acting in their best interest. %~\cite{Citation needed}.

Nonetheless, the second approach acknowledges that as the severity and persistence of discrepancies increase, individuals, even those with pro-social attitudes, may become less forgiving (Section~\ref{sec:minl}). Thus, the historical aspect of adverse effects becomes a critical factor. It is during extended periods of inconsistency that individuals may experience a diminishing willingness to tolerate disparities. This approach takes into account the temporal dimension of adverse effects, recognizing that extended discrepancies may erode the perception of fairness~\cite{jain1984quantitative}.

In our third approach, we examine a balance between the instantaneous adverse effect and the historical record of adverse effects (Section~\ref{sec:minfina+ul}). By combining these two dimensions, we aim to provide a tradeoff that accounts for both short-term variations and long-term consequences. This approach is particularly valuable in situations where CPS must navigate the delicate balance between immediate and lasting impact.

Building on the well-established literature on fair resource distribution, we acknowledge that fairness is not synonymous with equal resource allocation~\cite{jain1984quantitative}. Humans' perception of fairness is intrinsically tied to how they compare their resource distribution with that of others in the same system. In the fourth approach, we consider the collective perception of fairness among individuals who share the same CPS environment as a metric for formalizing F\emph{in}A (Section~\ref{sec:miny}). This approach recognizes that individuals may be more forgiving of discrepancies if they perceive that others are experiencing similar variations in resource allocation.

Lastly, our fifth approach introduces a tradeoff between human perception of fairness and a budget of allowable discrepancy between individual preferences and the applied CPS control action (Section~\ref{sec:miny+b}). By imposing limits on the magnitude of permissible discrepancies, this approach provides a tradeoff between accommodating individual preferences and ensuring collective fairness.

\subsection{Fairness-in-Adverse-Effect (F$in$A) Setup}
In all of these five approaches, we consider a society $S$ that consists of $N$ different individuals and a CPS that serves this society by providing shared control actions $a$ that may be tailored toward the preferences and behavior of some of those individuals. We assume that every individual $n \in N$ has a different set of $g$ preferred actions $A_n = \{a^n_1, a^n_2, \dots, a^n_g\}$ that can serve them better\footnote{While in our first definition $A_n$ we assume a discrete set of preferred actions, this can be extended to a continuous range of preferred actions.}. Suppose an adverse effect $v_n(a)$ on individual $n$ occurs due to the chosen control action~$a$. An initial mechanism to measure the \textbf{adverse effect} on each $v_n(a)$ is to assume that a preferred action $a_g^n \in A_n $ is inversely proportional to its adverse effects. That is, to measure the adverse effects of a chosen control action $v_n(a)$ on human $n$, we can use the distance between the set of preferred actions $\mathcal{A}_n$ and the chosen control action $a$, i.e., $ v_n(a)= \max_{a_g^n \in \mathcal{A}_n} \|a_g^n  - a \|$.

\subsection{{Approach \rom{1}: Formalizing the definition of F$in$A with instantaneous effect}}\label{sec:minfina}

In our first approach, we formalize F\emph{in}A by examining instantaneous adverse effects, delving into the immediate consequences of CPS control actions. 
 
Hence, an initial definition of F$in$A can be:

\vspace{-3mm}
\begin{equation}
    \text{F}in\text{A} = \min_{a\in \mathcal{A}} \| \mathbf{v}(a) - \frac{1}{N} \mathds{1}^T \mathbf{v}(a) \otimes \mathds{1}  \| + \lambda \| \frac{1}{N} \mathds{1}^T \mathbf{v}(a) \|,
    \label{eq:finA}
\end{equation}
where $\mathbf{v}(a) = [v_1(a), v_2(a), \dots, v_N(a)]^\intercal$ and $\mathcal{A} =  \bigcup_{n=1}^N A_n $. In other words, Equation~\eqref{eq:finA} aims to choose the control action $a$, out of all possible $A_n$, that minimizes the difference between the individual adverse effects on every individual $v_n$ and the average of the adverse effects across all individuals $\frac{1}{N} \sum_{n=1}^N v_n(a) = \frac{1}{N} \mathds{1}^T \mathbf{v}(a)$. Indeed, one trivial solution will be to increase the adverse effect on all individuals to achieve the same average. Therefore, the second term in Equation~\eqref{eq:finA} asks that the chosen control action also aims to minimize the average adverse effect.

\subsection{Approach \rom{2}: Formalizing the definition of F$in$A using long-term temporal variations in adverse effect}\label{sec:minl}

% \m{Explain  here the motivation}

% Motivated by the consequences of long-term adverse effects as motivated by social psychologists~\cite{????}. We formulated another definition for F$in$A that only considers the long-term effects.

In the second approach, the historical context of adverse effects plays a pivotal role. Prolonged instances of inconsistency are where individuals may exhibit a reduced willingness to endure disparities. This approach places a significant emphasis on the temporal dimension of adverse effects, acknowledging that extended discrepancies can undermine the perception of fairness~\cite{jain1984quantitative}.

We define a long-term adverse effect $\mathbf{v}_n$ by monitoring the adverse effect for every applied action $a$ over a time horizon $T$ on human $n$. 

\begin{align}\label{eq:longadverse}
\mathbf{v}_n = [v_n^0, v_n^1, \dots, v_n^{T-1}]^\intercal,
\end{align}

where $v_n^j$ represents the adverse effect occurred at time $j$ for human $n$. However, to focus more on the recent adverse effects, we assign different weights to $v_n^j$ and calculate an accumulated value $u_n^t$ that represents the current history of adverse effects on human $n$ from time $t-T$ till $t-1$.

\begin{align}\label{eq:utility}
u_n^t = \frac{1}{T} \sum_{j=0}^{T-1} \frac{j}{T} v_n^j, \text{ for } n = 1,2,\dots, N
\end{align}

The historical adverse effect on all $N$ individuals can be represented by:
\begin{align}\label{eq:historyadverse}
\mathbf{u} = [u_1^t, u_2^t, \dots, u_N^t]^\intercal 
\end{align}

We can then formally define F$in$A as: 

\begin{align}\label{eq:longtermform}
\begin{split}
\text{F}in\text{A} = &\min_{a \in \mathcal{A}}\ \mathcal{B} \\
s.t.\\
& \mathbf{v}(a) < \mathcal{B} - \mathbf{u},  %\forall n\\
\end{split}
\end{align}

In this equation, $\mathbf{v}(a) = [v_1(a), v_2(a), \dots, v_N(a)]^\intercal$ represents the current adverse effect of action $a$ on each individual. To elaborate, Equation~\eqref{eq:longtermform} contains $N$ constraints, with each constraint governing the current adverse effect $v_n(a)$ to remain below a specific budget $\mathcal{B}$. However, it's essential to note that this budget $\mathcal{B}$ is gauged on the historical adverse effects of each individual, denoted as $u_n^t$. In this context, F\emph{in}A tries to identify the minimum budget $\mathcal{B}$ that satisfies all N constraints. Consequently, if a human individual, say $n$, has a substantial historical adverse effect value, the constraint $\mathcal{B} - u_n^t$ will direct the optimization process toward finding an action $a$ that results in a minimal $v_n(a)$. This approach is designed to steer the optimization's focus towards individuals with higher historical adverse effect values, encouraging the selection of new actions that minimize their adverse effects.

% where $\mathbf{v}(a) = [v_1(a), v_2(a), \dots, v_N(a)]^T$ representing the current adverse effect of action $a$ on each individual. In other words, Equation~\ref{eq:longtermform} has $N$ constraints where each constraint bounds the current adverse effect $v_n(a)$ to be less than a particular budget $\mathcal{B}$. However, this budget $\mathcal{B}$ is gauged by the historical adverse effects of this individual $u_n^t$. In this case, F$in$A tries to find the minimum budget $\mathcal{B}$ that can satisfies all these $N$ constraints. Hence, if human $n$ has high value of historical adverse effects, this bound $\mathcal{B} - u_n^t$ will force the optimization to find an $a$ that causes small $v_n(a)$. The motivation is to push our optimization towards giving more focus to humans with more historical adverse effect value by choosing a new action $a$ that reduces their $v(a)$.  

\subsection{Approach \rom{3}: Formalizing the definition of F$in$A as a tradeoff between instantaneous and long-term adverse effect}\label{sec:minfina+ul}

% Within our third approach, we explore the tradeoff between the immediate adverse effect and the accumulated historical adverse effects. Accordingly, we need to append Equation~\eqref{eq:finA} with a term that depends on the accumulated historical adverse effect $u_n^t$ (Equation~\eqref{eq:utility}).  

% \begin{align}\label{eq:fina+ul}
% \begin{split}
%     \text{F}in\text{A} =  \min_{a\in \mathcal{A}, \mathbf{b}}\ &\alpha \Big( \| \mathbf{v}(a) - \frac{1}{G} \mathds{1}^T \mathbf{v}(a) \otimes \mathds{1}  \| +  \lambda \| \frac{1}{G} \mathds{1}^T \mathbf{v}(a) \|  \Big) \\ & + \beta \mathbf{u}^T \mathbf{b}, \\
%     s.t.\  & \mathbf{v}(a) < \mathbf{b}  %\forall n\\
% \end{split}
% \end{align}

% where $\mathbf{b} = [b_1, b_2, \dots, b_n]^T$ and $\mathbf{u} = [u_1^t, u_1^t,, \dots, u_n^t]^T$. In other words, we add a new optimization variable $b$ for every human $n$ that represents a budget for $v(a)$ we allow for every human $n$. However, these budgets $\mathbf{b}$ are weighted by the value of the long-term adverse effect $\mathbf{u}$. We add parameters $\alpha$ and $\beta$ to have a tradeoff between instant and long-term adverse effects.

In our third approach, we delve into the balance between immediate adverse effects and cumulative historical adverse effects. To achieve this, we augment Equation~\eqref{eq:finA} with a term that considers the historical adverse effects, denoted as $u_n^t$ as defined in Equation~\eqref{eq:utility}.

The extended formulation can be expressed as:

\begin{align}\label{eq:fina+ul}
\begin{split}
    \text{F}in\text{A} =  \min_{a\in \mathcal{A}, \mathbf{b}}\ &\alpha \Big( \| \mathbf{v}(a) - \frac{1}{G} \mathds{1}^T \mathbf{v}(a) \otimes \mathds{1}  \| +  \lambda \| \frac{1}{G} \mathds{1}^T \mathbf{v}(a) \|  \Big) \\ & + \beta \mathbf{u}^\intercal \mathbf{b}, \\
    s.t.\\
    & \mathbf{v}(a) < \mathbf{b}  %\forall n\\
\end{split}
\end{align}

In this formulation, $\mathbf{b} = [b_1, b_2, \dots, b_N]^\intercal$ and $\mathbf{u} = [u_1^t, u_2^t, \dots, u_N^t]^\intercal$. Essentially, the first term in Equation~\eqref{eq:fina+ul} is from the instantaneous adverse effect (Equation~\eqref{eq:finA}), and then we introduce $N$ new optimization variables, $b_n$, for each individual $n$, which represents the budget allocated for the adverse effect ($v_n(a)$) for each individual $n$. Importantly, these budgets $\mathbf{b}$ are weighted by the values of the long-term adverse effects $\mathbf{u}$. 

In other words, we have $N$ constraints for different budgets for adverse effects that are bounded for every individual  ($\mathbf{v}(a) < \mathbf{b}$). Accordingly, F$in$A needs to minimize these budgets $\mathbf{b}$ to minimize the overall adverse effects. However, the upper bound for these budgets is weighted by the accumulated historical adverse effects. Hence, the term  $\mathbf{u}^\intercal \mathbf{b}$ is appended to the definition of F$in$A.

To modulate the tradeoff between immediate and long-term adverse effects, we introduce parameters $\alpha$ and $\beta$.

\subsection{Approach \rom{4}: Formalization the definition of F$in$A using human perception of fairness}\label{sec:miny}

Drawing upon extensive research in the realm of equitable resource distribution, we recognize that fairness does not necessarily mean equal resource allocation~\cite{jain1984quantitative}. Instead, the perception of fairness in individuals is fundamentally linked to how they gauge their own resource distribution concerning that of their peers within the same system. In particular, fairness in this setup can be viewed generally as ``variances''~\cite{jain1984quantitative} of the ``utility'' shared by individuals. Hence, we exploit the notion of the coefficient of variation ($CoV$) of the utilities~\cite{jain1984quantitative}. In our setup, we define this utility for human $n$ at time $t$ as the temporal accumulated adverse effect caused by the control actions of the CPS in the shared environment, denoted by $u_n^t$ as expressed in Equation~\eqref{eq:utility}.

\begin{align}\label{eq:cov(u)}
 CoV_\mathbf{u} = \sqrt{\frac{1}{N-1} \sum_{n=1}^{N} \frac{(u_n - \bar{\mathbf{u}})^2}{\bar{\mathbf{u}}^2}}
\end{align}

In Equation~\eqref{eq:cov(u)}, $\bar{\mathbf{u}} = \frac{1}{N} \sum_{n=1}^N u_n^t$ represents the average utility (accumulated adverse effect at time $t$) of all $N$ humans. The system is said to be more fair if and only if the $CoV$ is smaller. The value of $CoV$ can be anywhere between $0$ and infinity. Hence, we use the fairness index ($FI$) transformation to have a value between $0$ and $1$ to be easily interpreted as a fairness percentage. In other words, if $FI$ is $1$, it means the system is $100\%$ fair, otherwise, if disparity increases between individuals, this $FI$ value will decrease~\cite{jain1984quantitative}.

\begin{align}\label{eq:fi(u)}
FI_\mathbf{u} =\frac{1}{1+CoV_\mathbf{u}^2}
\end{align}

Accordingly, we can define F$in$A to maximize the fairness index. The core idea is to minimize the reciprocal of this fairness index, represented as 
$y$ in the optimization.

\begin{align}\label{eq:fairindexform}
\begin{split}
&\text{F}in\text{A}=\min_{a \in \mathcal{A}}\ y\\
s.t. \\
& y \geq 1 + \frac{1}{N} \sum_{n=1}^{N} \left(\frac{|u_n - \bar{\mathbf{u}}|}{\bar{\mathbf{u}}}\right)^2\\
&\frac{|u_n - \bar{\mathbf{u}}|}{\bar{\mathbf{u}}} \leq \epsilon, \text{ for } n = 1, 2, \ldots, N. \\
\text{where:}\\
& u_n = u_n^t + v(a)
\end{split}
\end{align}

In this formulation, $u_n^t$ is a constant value that represents the accumulated history of the adverse effect up till time $t-1$ (Equation~\eqref{eq:utility}) before applying the new $a$ that will add the new adverse effect  $v(a)$ on human $n$. We also add the constraint $\frac{|u_n - \bar{\mathbf{u}}|}{\bar{\mathbf{u}}} \leq \epsilon, \text{ for } n = 1, 2, \ldots, N$ that sets a limit on how much each individual's adverse effect can deviate from the average adverse effect. The parameter $\epsilon$ defines the maximum allowable difference.

\subsection{Approach \rom{5}: Formalizing the definition of FinA using the human perception of fairness with a tradeoff for a budget for long-term adverse effect}~\label{sec:miny+b}

Lastly, our fifth approach introduces a tradeoff between human perception of fairness and a budget of allowable discrepancy between individual preferences and the applied CPS control action. In particular, we combine our definition of F$in$A in Equation~\eqref{eq:longtermform} and Equation~\eqref{eq:fairindexform} to provide a tradeoff between fairness index and adverse effect budget $\mathcal{B}$. 

\begin{align}
\begin{split}
&\text{F}in\text{A}=\min_{a \in \mathcal{A}}\ \alpha . y + \beta . \mathcal{B} \\
s.t.\\
& y \geq 1 + \frac{1}{N} \sum_{n=1}^{N} \left(\frac{|u_n - \bar{\mathbf{u}}|}{\bar{\mathbf{u}}}\right)^2\\
%& u_n < \mathcal{B}, \text{ for } n = 1, 2, \ldots, N.  \\
 & \mathbf{v}(a) < \mathcal{B} - \mathbf{u} \\
\text{where:}\\
& u_n = u_n^t + v(a)\\ 
%& v(a) < \mathcal{B} - u_n; \forall n \text{ This is the old calculated } u_n \text{from previous time step}
\end{split}
\end{align}

The $\alpha$ and $\beta$ weights allow us to adjust the tradeoff between fairness index and adverse effect budget.

%% file: 04_application_house.tex
\begin{table*}[ht]
    \centering
\begin{tabularx}{\linewidth}{XXX}
\includegraphics[width=\linewidth]{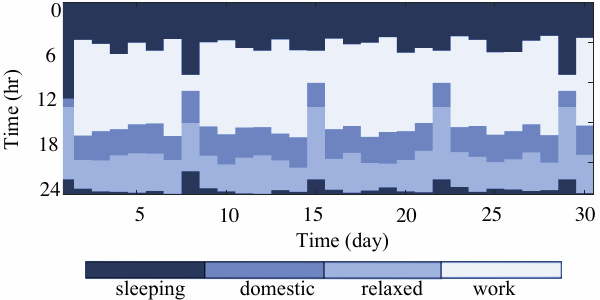}
    \captionof{figure}{Human 1 activity pattern.}\label{fig:act1}
& \includegraphics[width=\linewidth]{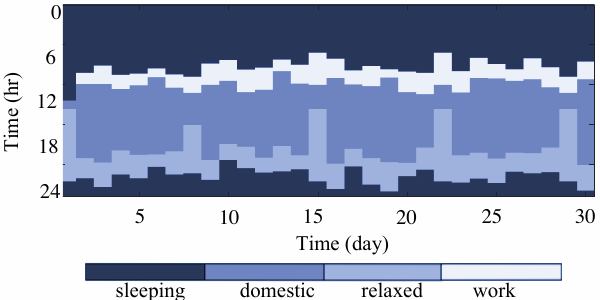}   
    \captionof{figure}{Human 2 activity pattern.}\label{fig:act2}    
& \includegraphics[width=\linewidth]{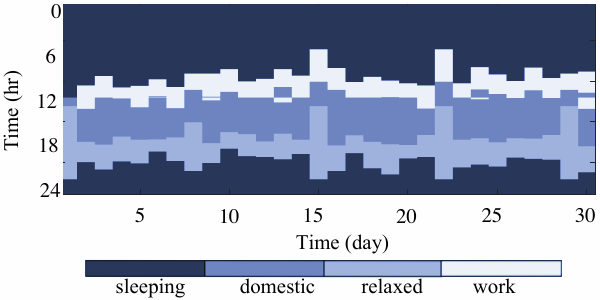}
     \captionof{figure}{Human 3 activity pattern.}\label{fig:act3} 
\end{tabularx}
%\caption{A table with figures}
%\vspace{-7mm}
\label{tab:humanactivityapp1}
\end{table*}

\section{Evaluation}\label{sec:evaluation}
We designed an HCPS application in the domain of smart house. Recent literature focuses on enhancing human satisfaction in smart heating, ventilation, and air conditioning (HVAC) systems by employing various techniques to adjust the set-point based on human activity and preferences~\cite{jung2017towards,elmalaki2021fair}. These HCPS systems consider the current state and individual preferences, such as body temperature changes during sleep or physical activity. %Monitoring human state, sleep cycle, and physical activity can be done using IoT edge devices~\cite{likamwa2013moodscope}. 
To evaluate different approaches of F$in$A in this application, we consider a setup where multiple humans share a house with a single HVAC system, and their activities determine individual HVAC set-point preferences. 

% \begin{table*}[!t]
%     \centering
% \begin{tabularx}{\linewidth}{XXX}
% \includegraphics[width=\linewidth]{figures/act1_modified.pdf}
%     \captionof{figure}{Human 1 activity pattern.}\label{fig:act1}
% & \includegraphics[width=\linewidth]{figures/act2_modified.pdf}   
%     \captionof{figure}{Human 2 activity pattern.}\label{fig:act2}    
% & \includegraphics[width=\linewidth]{figures/act3_modified.pdf}
%      \captionof{figure}{Human 3 activity pattern.}\label{fig:act3} 
% \end{tabularx}
% %\caption{A table with figures}
% \vspace{-7mm}
% \label{tab:humanactivityapp1}
% \end{table*}

\begin{figure*}[!t]
\centering
\includegraphics[width=0.95\linewidth]{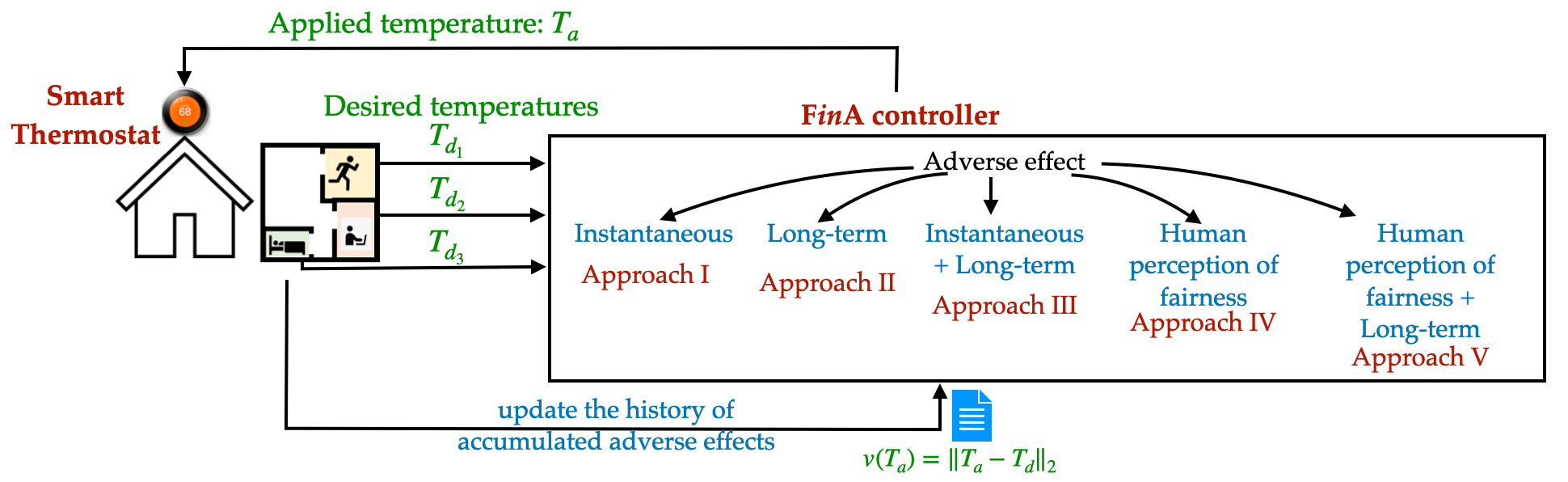}
\caption{A smart house with three humans. Each human has a different activity, which requires different desired indoor temperature setpoints $T_d$. An external controller running F$in$A selects the applied setpoint $T_a$ based on the calculations of instantaneous and long-term adverse effects.}\label{fig:appsetup}
\end{figure*}

We exploited recent work in the literature~\cite{elmalaki2021fair} that simulated a thermodynamic model of a house incorporating the house's shape %, window count, roof pitch angle, 
and insulation type. 
To regulate indoor temperature, a heater and a cooler with specific flow temperatures ($50^{\circ}c$ and $10^{\circ}c$) were employed. A thermostat maintained the indoor temperature within $2.5^{\circ}c$ around the desired set point. An external controller controls the setpoint running the optimization of F$in$A. A pictorial figure of the application setup is shown in Figure~\ref{fig:appsetup}.

We implemented our proposed five approaches using CVXPY, a Python-embedded modeling language for convex optimization problems~\cite{diamond2016cvxpy}. 

The human was modeled as a heat source, with heat flow dependent on the average exhale breath temperature ($EBT$) and the respiratory minute volume ($RMV$). %  and metabolic rate~\cite{metactivity}. 
These parameters depend on human activity~\cite{carrollpulmonary2007}. We simulated three humans with four activities: sleeping, relaxing, medium domestic work, and working from home. Randomness was introduced by allowing multiple activity choices during the same time slot. %For simplicity, activities with minor $RMV$ differences were grouped together.
The different activity schedules depicted in Figures \ref{fig:act1}, \ref{fig:act2}, and \ref{fig:act3}.  %Age, sex, and time of day were not considered. 
The humans were simulated in separate rooms as seen in Figure~\ref{fig:appsetup}, each exhibiting unique behavioral patterns: (1) $h_1$ followed an organized and repetitive weekly routine, (2) $h_3$ had a more random and unpredictable life pattern, and (3) $h_2$ displayed intermediate randomness, alternating between sleeping, being away from home, domestic activities, and relaxation.  The Mathworks thermal house model was extended to include a cooling system and a human model\footnote{While more complex simulators like EnergyPlus~\cite{gerber2014energyplus} exist, considering energy consumption and electric loads, we opted for a simpler model to assess F$in$A.}. 

The desired preferred action (temperature setpoint) per human $a_g^n = T_n$, for $n =1,2,$ and $3$, can be obtained through fixed policy configuration. We exploit existing approaches~\cite{taherisadr2023adaparl} for estimating the desired HVAC setpoint based on activity and thermal comfort. The desired setpoints for the considered activities are domestic activity ($72^\circ$F), relaxed activity ($77^\circ$F), sleeping ($62^\circ$F), and work from home ($67^\circ$F). These setpoints aim to enhance thermal comfort~\cite{fanger1970thermal}.%,handbook2009american}. %Thermal comfort is assessed using Prediction Mean Vote (PMV) on a scale from very cold ($-3$) to very hot ($+3$)\cite{fanger1970thermal}. Optimal indoor thermal comfort falls within the recommended range of $[-0.5, 0.5]$, as per the ISO standard ASHRAE $55$~\cite{handbook2009american}. 

\begin{table*}[!t]
    \centering
\begin{tabularx}{\textwidth} {p{0.005\textwidth} || p{0.23\textwidth} | p{0.23\textwidth} | p{0.23\textwidth} | p{0.23\textwidth}}

% & Accumulated adverse effect(u) & Histogram of absolute temperature difference & Satisfaction sample rate & Histogram of satisfaction sample rate
& \multicolumn{1}{c|}{Accumulated adverse effect($\mathbf{u}$) } & \multicolumn{1}{c|}{Hist. temperature difference $|T_{diff}|$} & \multicolumn{1}{c|}{Satisfaction rate (SR) \%} & \multicolumn{1}{c}{Hist. satisfaction rate (SR)} \\\hline\hline
%%%%%%%% Approach 1
\raisebox{3ex}{\rotatebox{90}{Approach \rom{1}}} &  \includegraphics[height=3cm,width=0.25\textwidth]{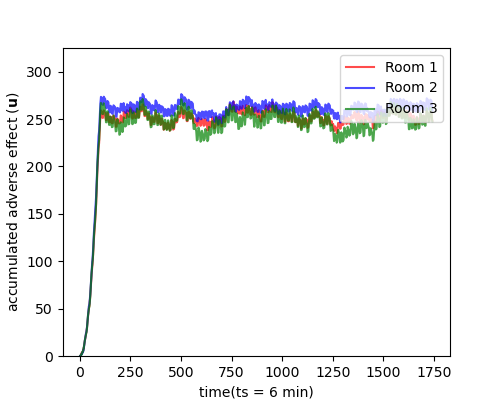}
    \label{fig:fina_u_all}  
    & \includegraphics[height=3cm,width=0.25\textwidth]{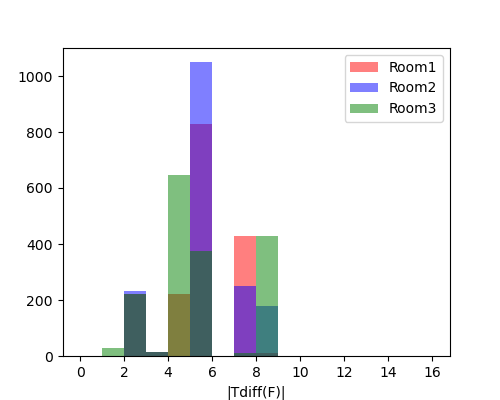}
     \label{fig:fina_tdiff_abs_hist} 
     & \includegraphics[height=3cm,width=0.25\textwidth]{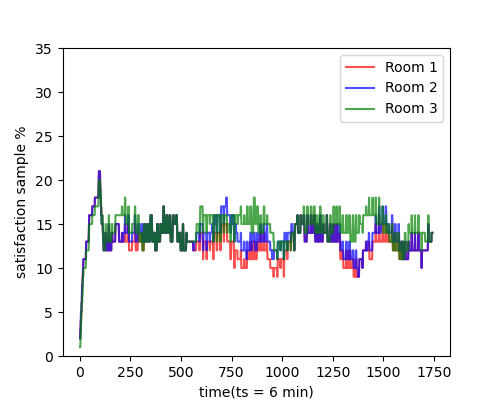}
     \label{fig:fina_sat_sample_number}
     & \includegraphics[height=3cm,width=0.25\textwidth]{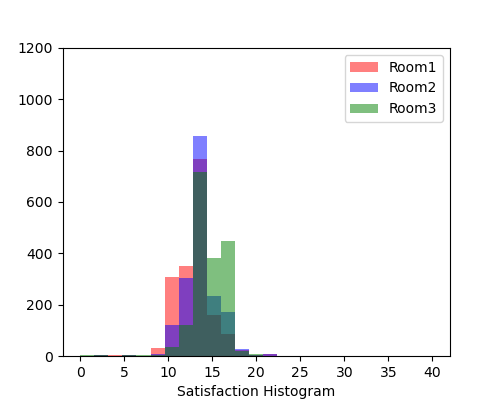}
     \label{fig:fina_sat_sample_number_hist} \\\hline
%%%%%%%% Approach 2     
\raisebox{3ex}{\rotatebox{90}{Approach \rom{2}}} & \includegraphics[height=3cm,width=0.25\textwidth]{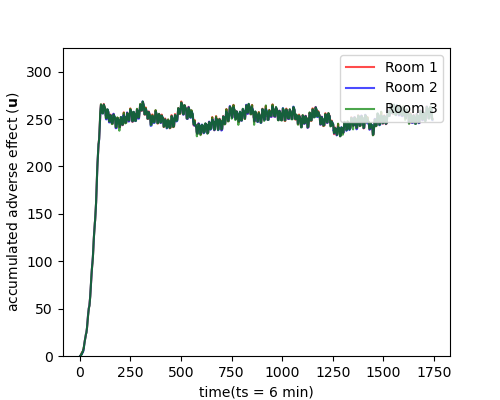}
    \label{fig:l_u_all}  
    & \includegraphics[height=3cm,width=0.25\textwidth]{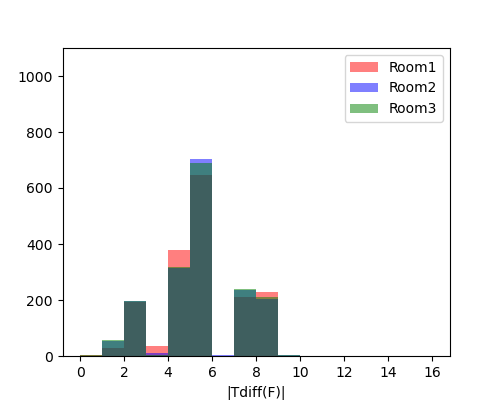}
     \label{fig:l_tdiff_abs_hist} 
     & \includegraphics[height=3cm,width=0.25\textwidth]{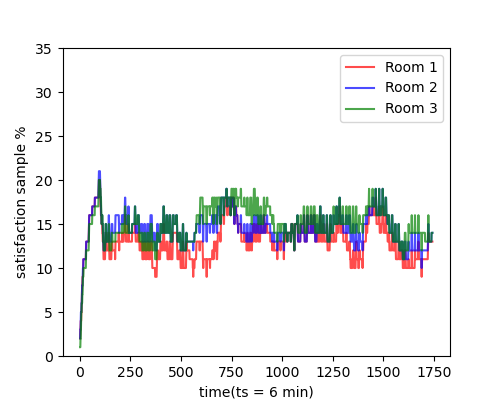}
     \label{fig:l_sat_sample_number} 
     & \includegraphics[height=3cm,width=0.25\textwidth]{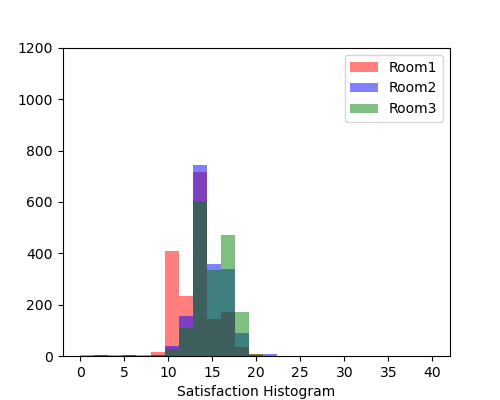}
     \label{fig:l_sat_sample_number_hist} \\\hline
%%%%%%%% Approach 3     
\raisebox{3ex}{\rotatebox{90}{Approach \rom{3}}} &
\includegraphics[height=3cm,width=0.25\textwidth]{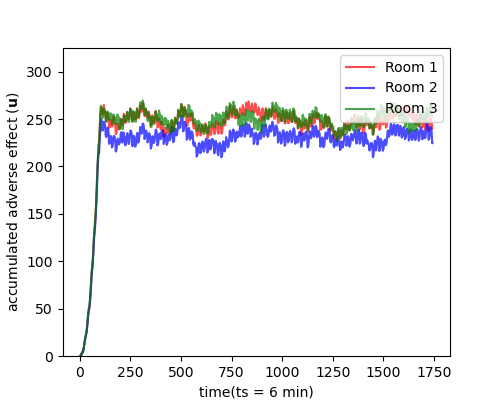}
    \label{fig:fina_ul_u_all}  
    & \includegraphics[height=3cm,width=0.25\textwidth]{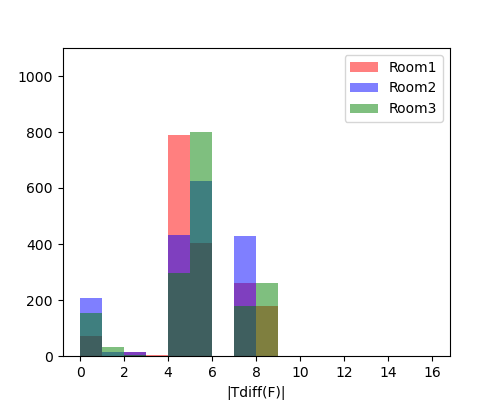}
     \label{fig:fina_ul_tdiff_abs_hist} 
     & \includegraphics[height=3cm,width=0.25\textwidth]{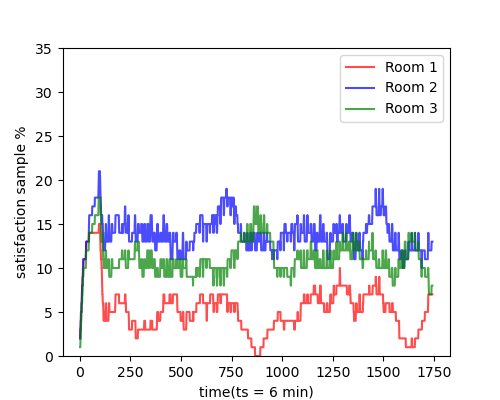}
     \label{fig:fina_ul_sat_sample_number} 
     & \includegraphics[height=3cm,width=0.25\textwidth]{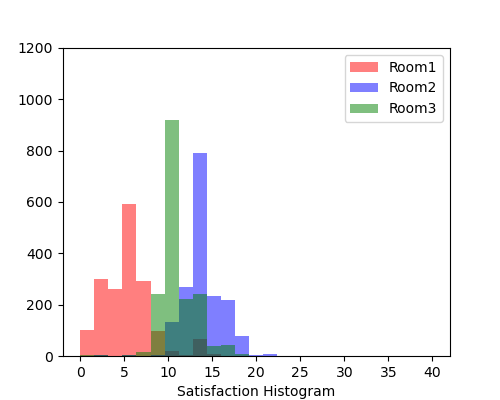}
     \label{fig:fina_ul_sat_sample_number_hist} \\\hline
%%%%%%%% Approach 4 
\raisebox{3ex}{\rotatebox{90}{Approach \rom{4}}} &\includegraphics[height=3cm,width=0.25\textwidth]{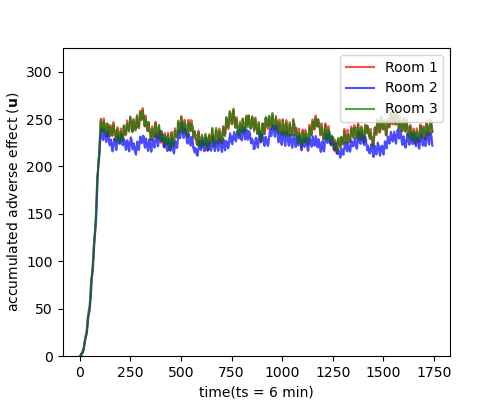}
    \label{fig:y_u_all}  
    & \includegraphics[height=3cm,width=0.25\textwidth]{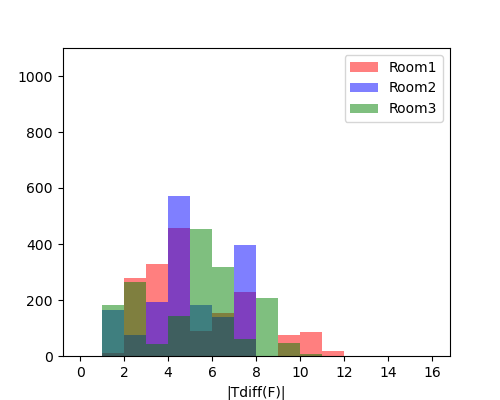}
     \label{fig:y_tdiff_abs_hist} 
     & \includegraphics[height=3cm,width=0.25\textwidth]{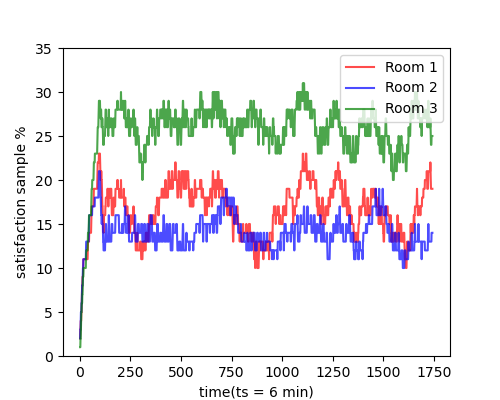}
     \label{fig:y_sat_sample_number} 
     & \includegraphics[height=3cm,width=0.25\textwidth]{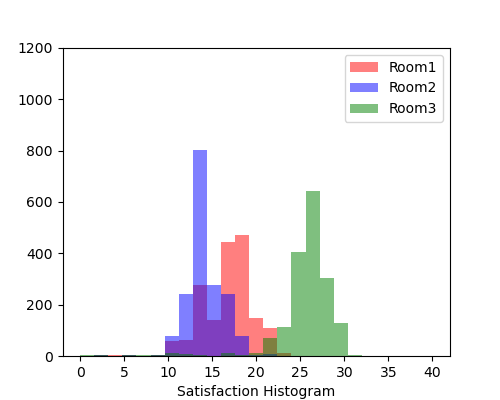}
     \label{fig:y_sat_sample_number_hist} \\\hline
%%%%%%%% Approach 5
\raisebox{3ex}{\rotatebox{90}{Approach \rom{5}}} & \includegraphics[height=3cm,width=0.25\textwidth]{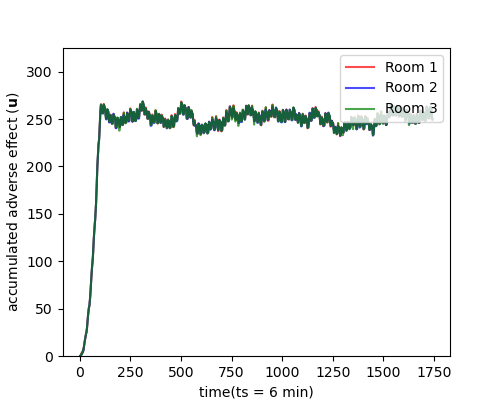}
    \label{fig:y_b_u_all}  
    & \includegraphics[height=3cm,width=0.25\textwidth]{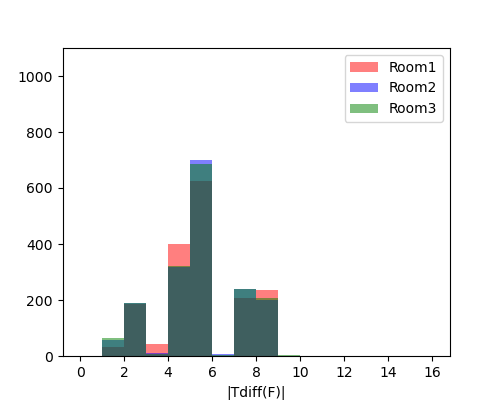}
     \label{fig:y_b_tdiff_abs_hist} 
     & \includegraphics[height=3cm,width=0.25\textwidth]{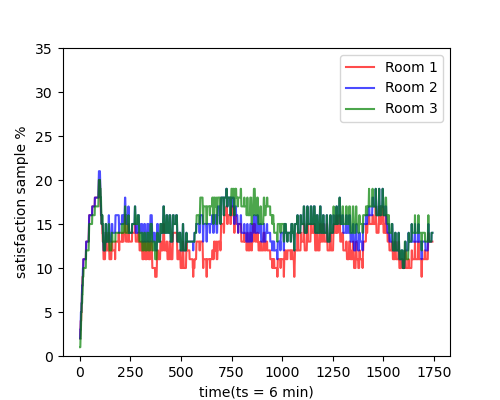}
     \label{fig:y_b_sat_sample_number} 
     & \includegraphics[height=3cm,width=0.25\textwidth]{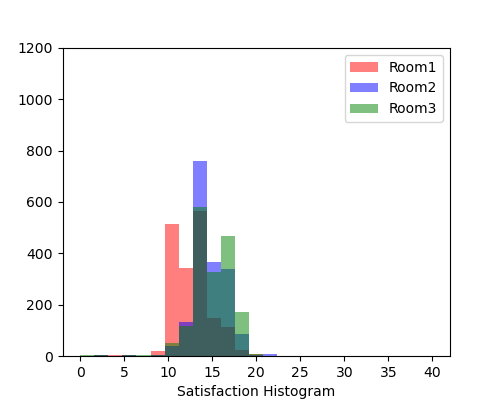}
     \label{fig:y_b_sat_sample_number_hist} \\\hline
%%%%%%%% Average   
\raisebox{3ex}{\rotatebox{90}{Mean}}
     & \includegraphics[height=3cm,width=0.25\textwidth]{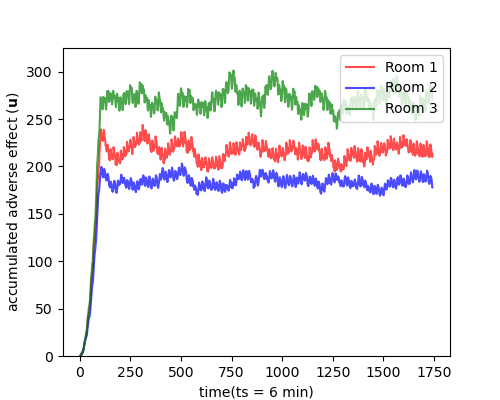}
    \label{fig:mean_u_all}  
    & \includegraphics[height=3cm,width=0.25\textwidth]{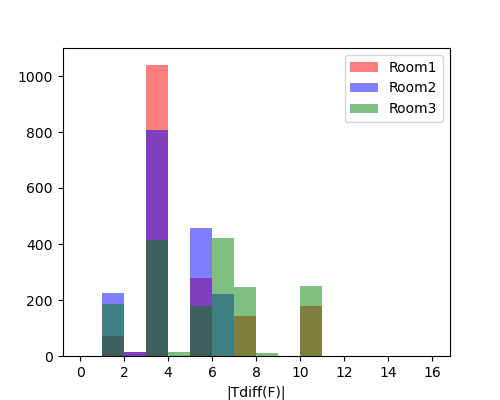}
     \label{fig:mean_tdiff_abs_hist} 
     & \includegraphics[height=3cm,width=0.25\textwidth]{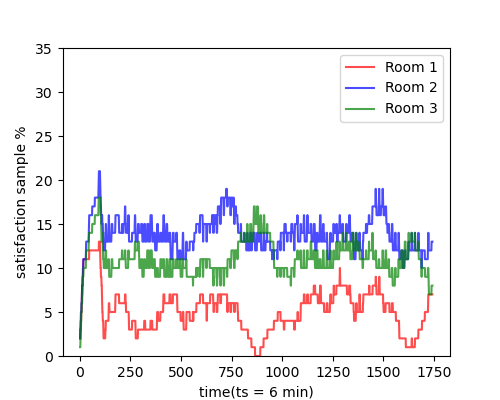}
     \label{fig:mean_sat_sample_number} 
     & \includegraphics[height=3cm,width=0.25\textwidth]{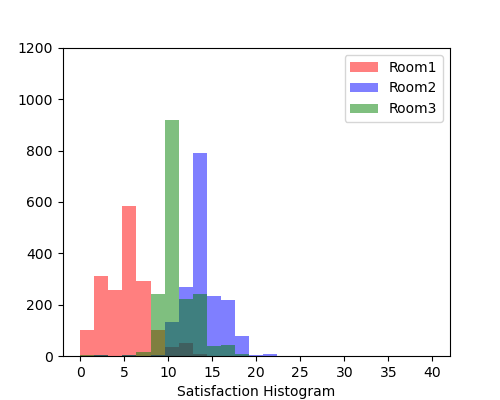}
     \label{fig:mean_sat_sample_number_hist} \\\hline
%%%%%%%% RR  
\raisebox{3ex}{\rotatebox{90}{Round Robin}}
     & \includegraphics[height=3cm,width=0.25\textwidth]{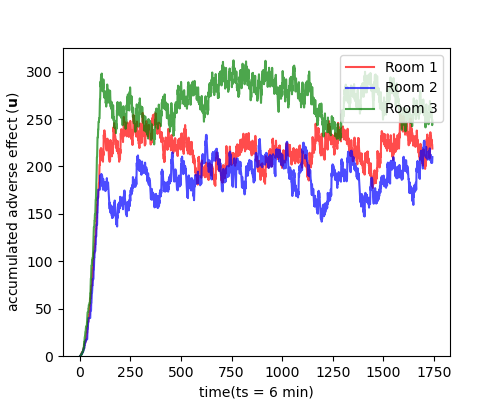}
    \label{fig:RR_u_all}  
    & \includegraphics[height=3cm,width=0.25\textwidth]{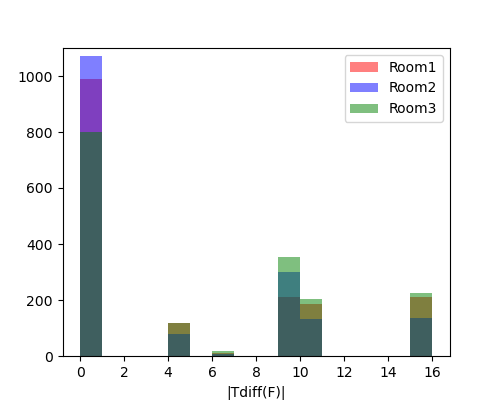}
     \label{fig:RR_tdiff_abs_hist} 
     & \includegraphics[height=3cm,width=0.25\textwidth]{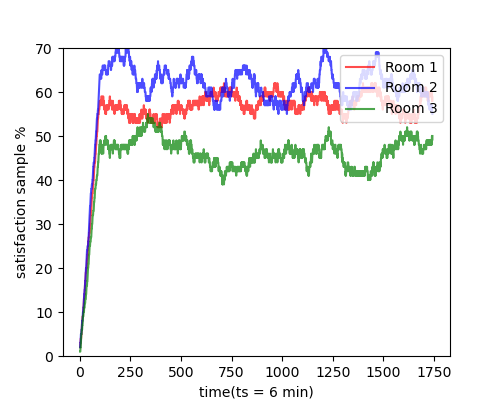}
     \label{fig:RR_sat_sample_number} 
     & \includegraphics[height=3cm,width=0.25\textwidth]{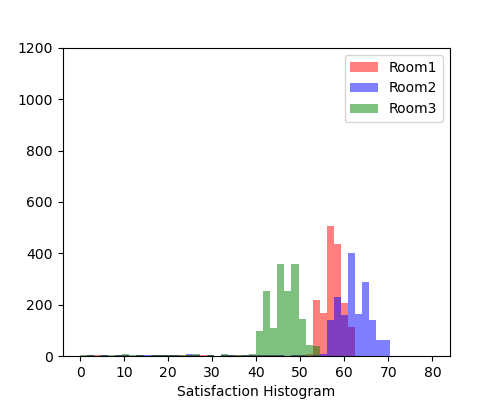}
     \label{fig:RR_sat_sample_number_hist} \\\hline
\end{tabularx}
\caption{Comparison between all the different five approaches of F$in$A, mean approach, and Round Robin}
\label{tbl: u_and_sat}
\end{table*}

%%%%%% FI AND CoV TABLE %%%%%%%
\begin{table*}[!t]
    \centering
\begin{tabularx}{\textwidth} {p{0.005\textwidth} || p{0.23\textwidth} | p{0.23\textwidth} | p{0.23\textwidth} | p{0.23\textwidth}}

% & Accumulated adverse effect(u) & Histogram of absolute temperature difference & Satisfaction sample rate & Histogram of satisfaction sample rate
& \multicolumn{1}{c|}{$FI_\mathbf{u}$ } & \multicolumn{1}{c|}{$CoV_\mathbf{u}$} & \multicolumn{1}{c|}{$FI_\mathbf{SR}$} & \multicolumn{1}{c}{$CoV_\mathbf{SR}$} \\\hline\hline
%%%%%%%% Approach 1
\raisebox{3ex}{\rotatebox{90}{Approach \rom{1}}} &  \includegraphics[height=3cm,width=0.25\textwidth]{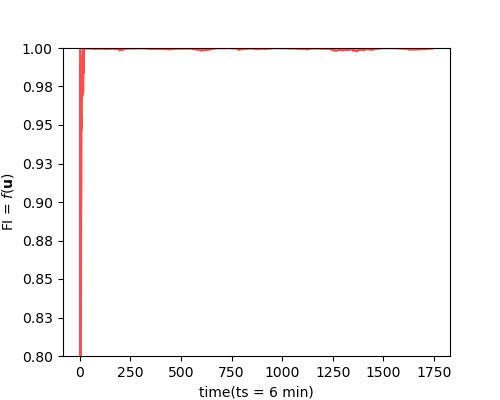}
    \label{fig:fina_u_FI}  
    & \includegraphics[height=3cm,width=0.25\textwidth]{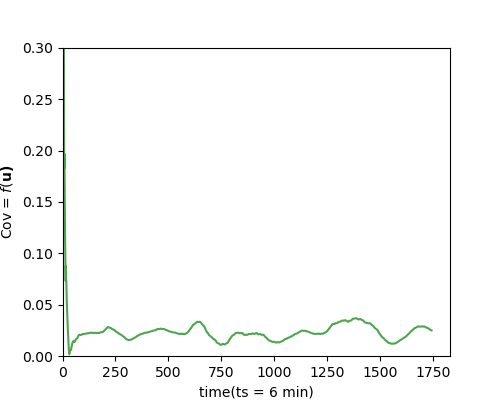}
     \label{fig:fina_u_CV} 
     & \includegraphics[height=3cm,width=0.25\textwidth]{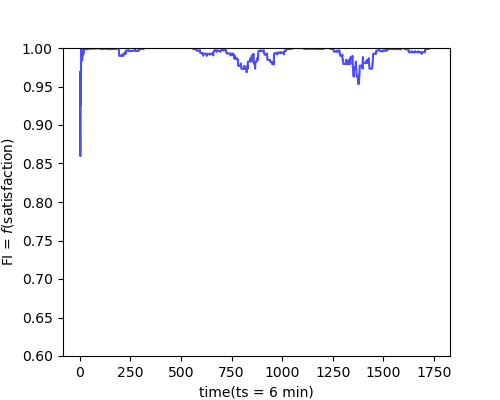}
     \label{fig:fina_satisfaction_FI}
     & \includegraphics[height=3cm,width=0.25\textwidth]{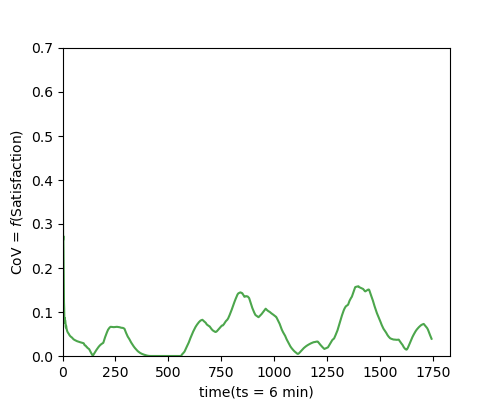}
     \label{fig:fina_sat_CV} \\\hline
%%%%%%%% Approach 2     
\raisebox{3ex}{\rotatebox{90}{Approach \rom{2}}} & \includegraphics[height=3cm,width=0.25\textwidth]{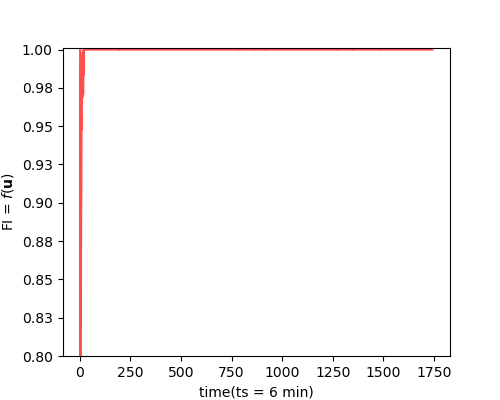}
    \label{fig:l_u_FI}  
    & \includegraphics[height=3cm,width=0.25\textwidth]{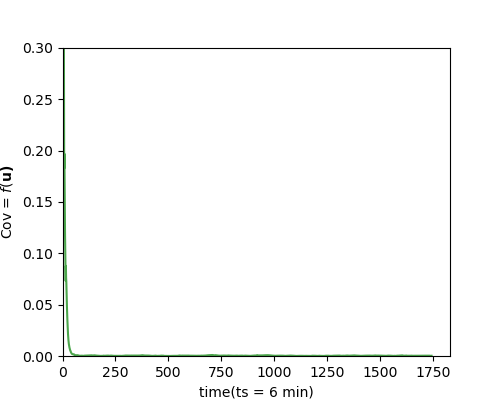}
     \label{fig:l_u_CV} 
     & \includegraphics[height=3cm,width=0.25\textwidth]{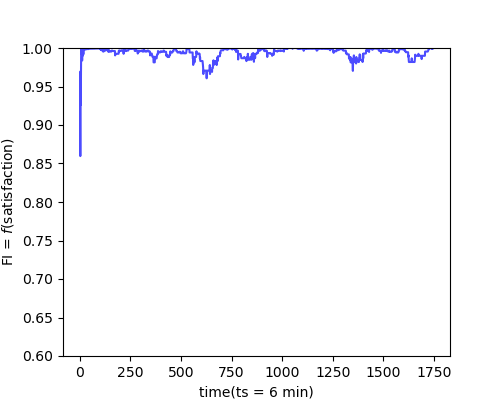}
     \label{fig:l_satisfaction_FI} 
     & \includegraphics[height=3cm,width=0.25\textwidth]{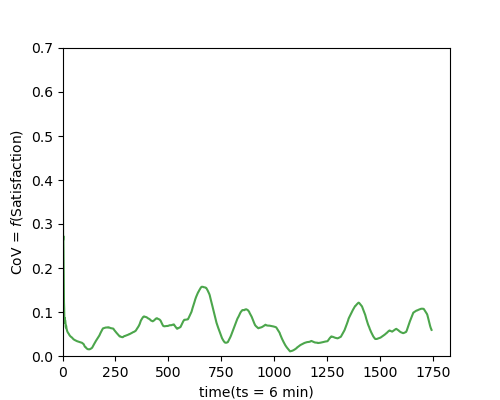}
     \label{fig:l_sat_CV} \\\hline
%%%%%%%% Approach 3     
\raisebox{3ex}{\rotatebox{90}{Approach \rom{3}}} &
\includegraphics[height=3cm,width=0.25\textwidth]{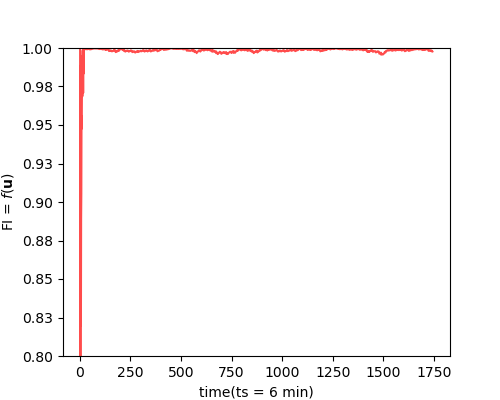}
    \label{fig:fina_ul_u_FI}  
    & \includegraphics[height=3cm,width=0.25\textwidth]{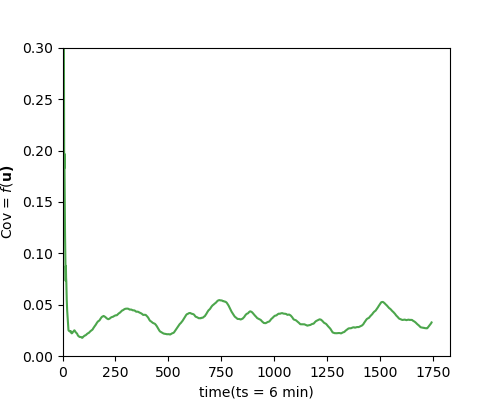}
     \label{fig:fina_ul_u_CV} 
     & \includegraphics[height=3cm,width=0.25\textwidth]{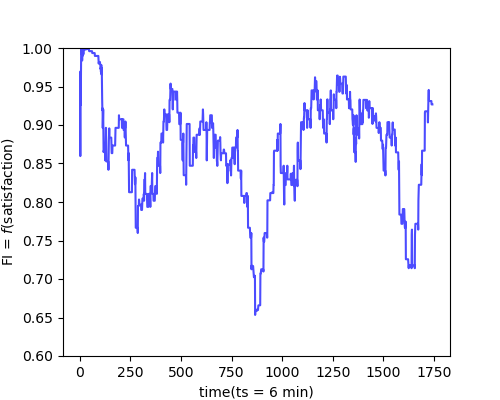}
     \label{fig:fina_ul_satisfaction_FI} 
     & \includegraphics[height=3cm,width=0.25\textwidth]{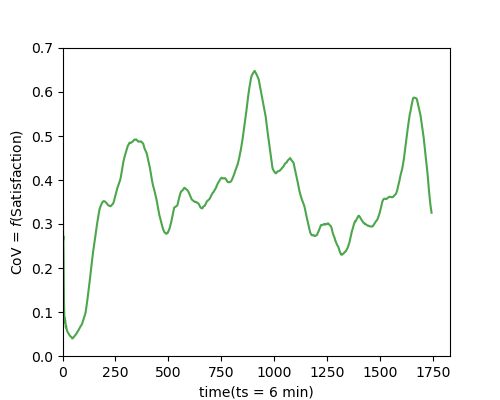}
     \label{fig:fina_ul_sat_CV} \\\hline
%%%%%%%% Approach 4 
\raisebox{3ex}{\rotatebox{90}{Approach \rom{4}}} &\includegraphics[height=3cm,width=0.25\textwidth]{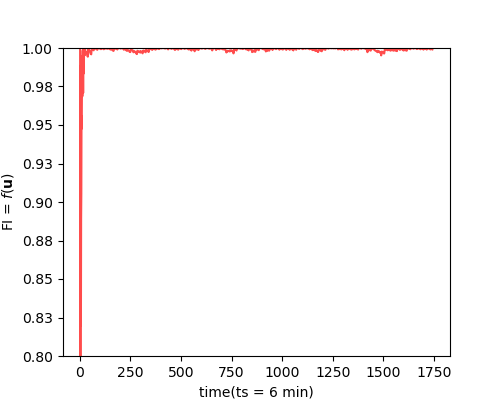}
    \label{fig:y_u_FI}  
    & \includegraphics[height=3cm,width=0.25\textwidth]{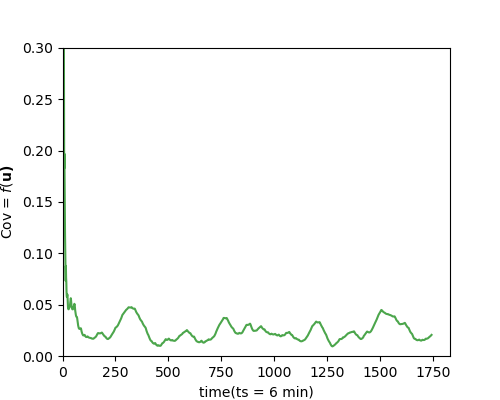}
     \label{fig:y_u_CV} 
     & \includegraphics[height=3cm,width=0.25\textwidth]{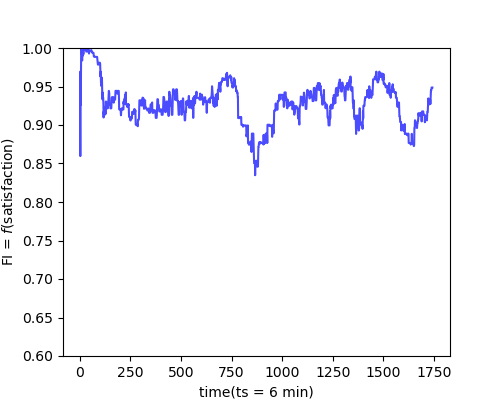}
     \label{fig:y_satisfaction_FI} 
     & \includegraphics[height=3cm,width=0.25\textwidth]{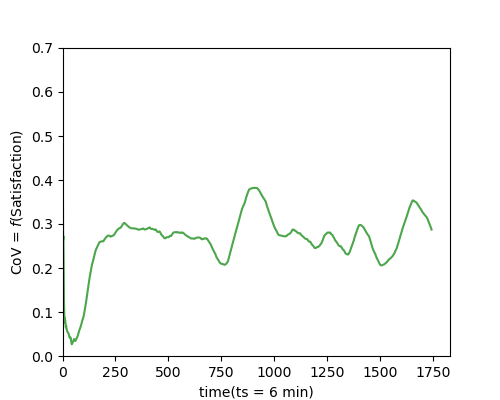}
     \label{fig:y_sat_CV} \\\hline
%%%%%%%% Approach 5
\raisebox{3ex}{\rotatebox{90}{Approach \rom{5}}} & \includegraphics[height=3cm,width=0.25\textwidth]{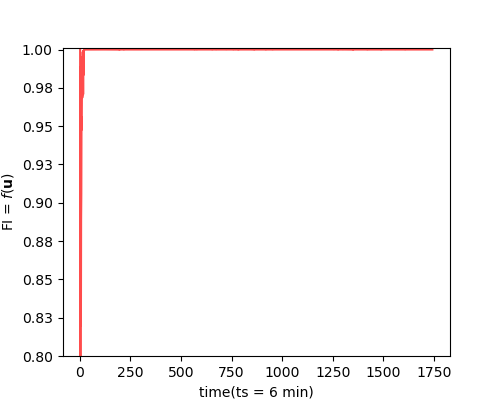}
    \label{fig:y_b_u_FI}  
    & \includegraphics[height=3cm,width=0.25\textwidth]{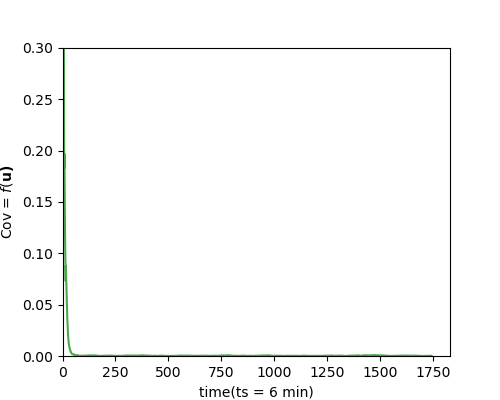}
     \label{fig:y_b_u_CV} 
     & \includegraphics[height=3cm,width=0.25\textwidth]{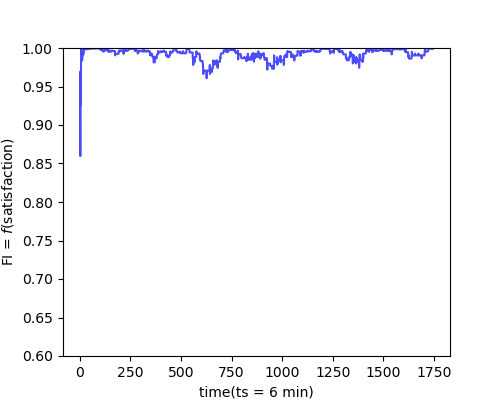}
     \label{fig:y_b_satisfaction_FI} 
     & \includegraphics[height=3cm,width=0.25\textwidth]{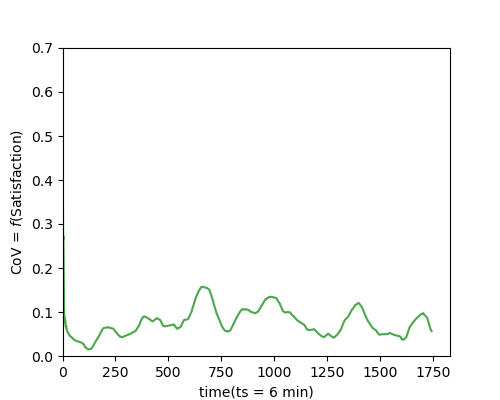}
     \label{fig:y_b_sat_CV} \\\hline
%%%%%%%% Average   
\raisebox{3ex}{\rotatebox{90}{Mean}}
     & \includegraphics[height=3cm,width=0.25\textwidth]{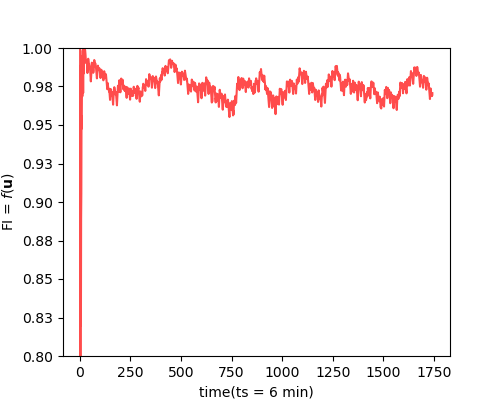}
    \label{fig:mean_u_FI}  
    & \includegraphics[height=3cm,width=0.25\textwidth]{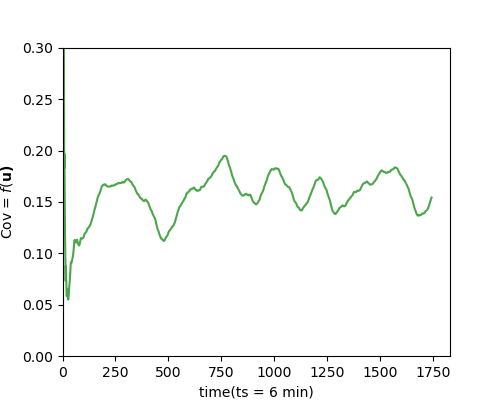}
     \label{fig:mean_u_CV} 
     & \includegraphics[height=3cm,width=0.25\textwidth]{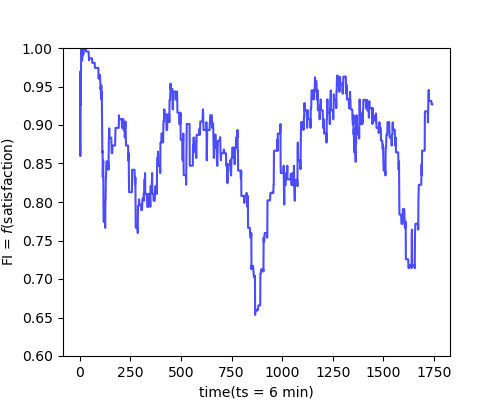}
     \label{fig:mean_satisfaction_FI} 
     & \includegraphics[height=3cm,width=0.25\textwidth]{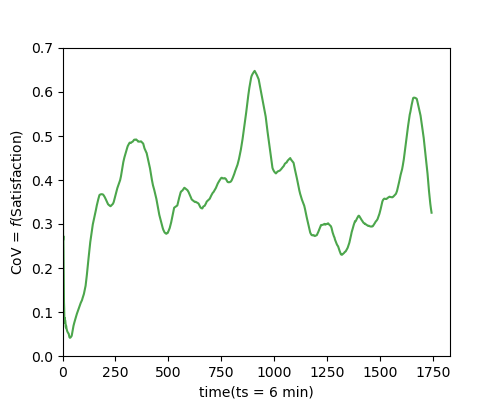}
     \label{fig:mean_sat_CV} \\\hline
%%%%%%%% RR  
\raisebox{3ex}{\rotatebox{90}{Round Robin}}
     & \includegraphics[height=3cm,width=0.25\textwidth]{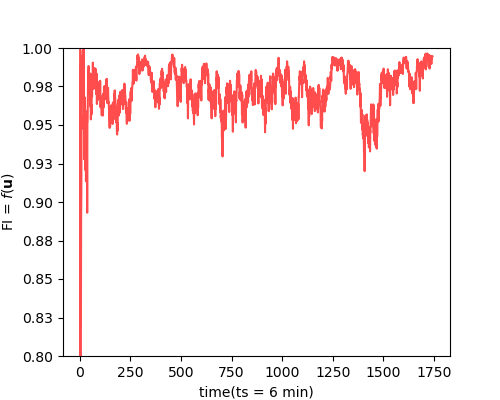}
    \label{fig:RR_u_FI}  
    & \includegraphics[height=3cm,width=0.25\textwidth]{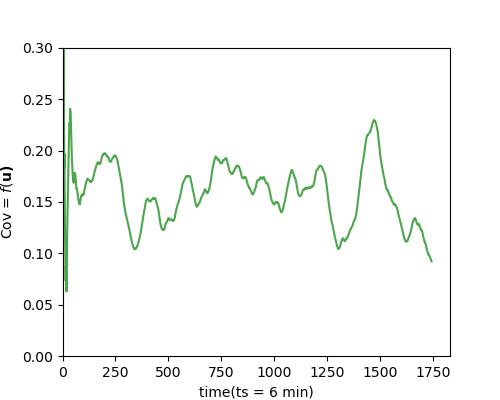}
     \label{fig:RR_u_CV} 
     & \includegraphics[height=3cm,width=0.25\textwidth]{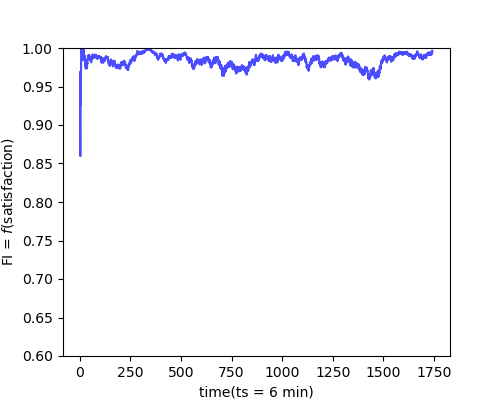}
     \label{fig:RR_satisfaction_FI} 
     & \includegraphics[height=3cm,width=0.25\textwidth]{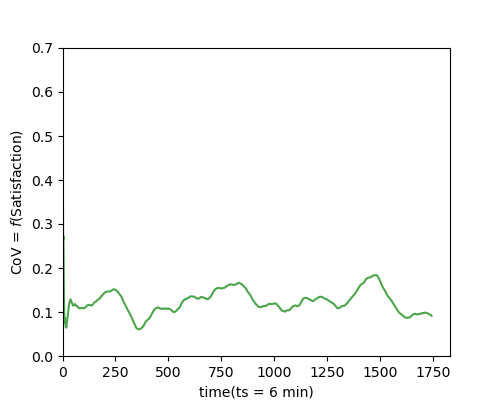}
     \label{fig:RR_sat_CV} \\\hline
\end{tabularx}
\caption{Comparison between all the different five approaches of F$in$A, mean approach, and Round Robin}
\label{tbl: FI_and_CoV}
\end{table*}

\subsection{Experiment setup}
In this application, we used the difference between the desired temperature ($T_d$) and the applied temperature ($T_a$) as a measure of the adverse effect: \[v(a) = \|T_a-T_d\|_2, \text{ where } T_a \in [60-80]^\circ F\]

We use the most recent $100$ samples for our history of adverse effects for the three humans $\mathbf{v}_1, \mathbf{v}_1$ and $\mathbf{v}_3$ (as explained in Equation~\eqref{eq:longadverse}) with sampling time $t_s = 6$ min. Hence, every $t_s$, we compute $\mathbf{u} = [u_1, u_2, u_3]^\intercal$ for the three humans (as explained in Equation~\eqref{eq:historyadverse}), where $u_n = \frac{1}{100}\sum_{j=0}^{99} \frac{j}{100} v_n^j$, and $v_n^j = \| T_a^j - T_{d_1}^j\|_2$ for $n=1,2,$ and $3$. 

The simulation was executed using a total of 3,000 samples, roughly equivalent to approximately 12 days. This extended duration allowed us to accurately capture changes in the behavioral patterns of the three individuals. We only ran F$in$A  when the desired temperatures of these three individuals were not identical.

We consider that the human is satisfied if the $T_a$ is within $2.5^\circ F$ difference from the desired temperature. We measure the satisfaction rate by considering the $100$ sample window in $\mathbf{v}_n$. Hence, the satisfaction rate ($SR$) for human $n$ is computed every $t_s$ computed as: \[  SR_n = \sum_{j=0}^{99} \mathds{1}(v(a) \leq 2.5), \text{ for all } v(a) \in \mathbf{v}_n  \]

Hence, the $SR$ can give us a measure in $\%$ since the total number of samples we consider is $100$. We use this SR to compute $CoV$ and $FI$ as a function of the $SR$ similar to Equations~\eqref{eq:cov(u)} and ~\eqref{eq:fi(u)} respectively. 
\begin{align}\label{eq:cov(sr)}
 CoV_\mathbf{SR} = \sqrt{\frac{1}{N-1} \sum_{n=1}^{N} \frac{(SR_n - \overline{\mathbf{SR}})^2}{\overline{\mathbf{SR}}^2}}, 
\end{align}

where $N=3$ and $\overline{\mathbf{SR}} = [SR_1, SR_2, SR_3]^\intercal$ computed for the three humans every $t_s$. 

Similarly, we consider the $FI_\mathbf{SR}$ as follows: 

\begin{align}\label{eq:fi(sr)}
FI_\mathbf{SR} =\frac{1}{1+CoV_\mathbf{SR}^2}
\end{align}

Furthermore, we compared the five proposed approaches for F$in$A with two more approaches: 
\begin{itemize}
\item \textbf{Mean approach:} In this case, the applied temperature $T_a$ is the mean of the desired temperature from the three humans. 
\item \textbf{Round Robin}: In this case, the applied temperature $T_a$ is selected in rotation between the desired temperature from the three humans. 
\item \textbf{FaiRIoT ~\cite{elmalaki2021fair}}: We compare with the state-of-the-art FaiRIoT, which uses hierarchical reinforcement learning to assign weights to the desired actions to compute the applied action. Hence, $T_a = \sum_{n=1}^3 w_n T_{d_n}$. 
\end{itemize}

In all of the experiments, we set the tradeoff parameters $\alpha=\beta=0.5$.

\subsection{Results}
%% explain the table 1 figures
We plot in Table~\ref{tbl: u_and_sat}, the accumulated adverse effect ($\mathbf{u} = [u_1, u_2, u_3]^\intercal$), the histogram of the absolute temperature difference between $|T_{diff}| = |T_a - T_d|$, the satisfaction rate ($SR$), and the histogram of the satisfaction rate ($SR$), across all approaches for the three individuals in three rooms.
% In Table~\ref{tbl: u_and_sat}, we show the accumulated adverse effect figure shows the summation of the latest $100$ samples' adverse effect. Histogram figure of absolute temperature difference calculates the distribution of the difference between applied temperature and desired temperature. Satisfaction sample rate figure plots the percentage of samples which the difference between applied temperature and desired temperature is less than threshold $2.5^{\circ}F$ in the latest $100$ samples. Histogram figure of satisfaction indicates the distribution of satisfaction sample rate. 

%explain accumulated adverse effect(u)
First column in Table~\ref{tbl: u_and_sat} compares the differences in the individual's adverse effect ($\mathbf{u} = [u_1, u_2, u_3]^\intercal$) across all approaches. Approach \rom{2} and \rom{5} show the smallest difference which is also reflected in average $CoV_\mathbf{u}$ in Table~\ref{tbl:stats}. Approach \rom{4} has a higher $CoV_\mathbf{u}$ (0.027) but it can bound $\mathbf{u}$ within a smaller value compared with other approaches observed in Table~\ref{tbl: u_and_sat}.

\begin{table}[H]
 % \small
  \centering
  \begin{tabularx}{0.99\columnwidth} {|p{0.09\columnwidth}|p{0.1\columnwidth}|p{0.1\columnwidth}|p{0.1\columnwidth}|p{0.1\columnwidth}|p{0.1\columnwidth}|p{0.1\columnwidth}|}
    \hline
 & $|T_{diff}|$ overlap\% & $SR$ JSD & $Avg.$ $FI_\mathbf{u}$ & $Avg.$ $CoV_\mathbf{u}$ & $Avg.$ $FI_\textbf{SR}$ & $Avg.$ $CoV_\textbf{SR}$ \\
\midrule
%Approach 1
\textbf{Appr. \rom{1}} & 22.4\% & 0.086 & 0.998 & 0.026 & \textbf{0.994} & \textbf{0.057}  \\
\hline
%Approach 2
\textbf{Appr. \rom{2}} & \textbf{86.5\%} & \textbf{0.010} & \textbf{0.999} & \textbf{0.004} & \textbf{0.994} & 0.066 \\
\hline
%Approach 3
\textbf{Appr. \rom{3}}& 37.6\% & 0.639 & 0.998 & 0.038 & 0.870 & 0.365  \\
\hline
%Approach 4
\textbf{Appr. \rom{4}} & 19.2\% & 0.659 & 0.998 & 0.027 & 0.929 & 0.624  \\
\hline
%Approach 5
\textbf{Appr. \rom{5}} & 83.4\% & 0.139 & \textbf{0.999} & \textbf{0.004} & 0.992 & 0.077 \\
\hline
%Mean
\textbf{Mean} & 24.8\% & 0.648 & 0.974 & 0.157 & 0.868 & 0.370  \\
\hline
%Average
\textbf{RR} & 68.4\% & 0.723 & 0.973 & 0.160 & 0.984 & 0.124  \\
\hline
\end{tabularx}
\caption{Comparison of the overlap area percentage, Satisfaction JSD, and average Fairness Index ($FI$) and the average coefficient of variation ($CoV$) of adverse effect($\textbf{u}$) and satisfaction(\textbf{SR}), respectively. }%\vspace{-4mm}
  \label{tbl:stats}
\end{table}

%explain Tdiff hist
We compare the distribution of $|T_{diff}|$ across all the approaches in  Table~\ref{tbl: u_and_sat} second column. Approach \rom{2} has the highest overlap percentage $86.5\%$ as calculated in Table~\ref{tbl:stats} which indicates that this approach can make all $3$ rooms have a more similar experience compared with other approaches. Round robin (RR) has a large overlap percentage due to the fact that each room can have a $|T_{diff}|=0$ on its turn in the round. However, RR will result in significant $|T_{diff}|$, which is larger than $10^{\circ}F$ in a notable number of the samples.

%explain satisfaction rate and hist of SR
Table~\ref{tbl: u_and_sat} third and fourth columns present the satisfaction rate ($SR$) across all approaches. We report the Jensen-Shannon Divergence (JSD) of the histogram for $SR$ in Table~\ref{tbl:stats}\footnote{The Jensen–Shannon divergence is a method of measuring the similarity between two probability distributions. The JSD is symmetric and always non-negative, with a value of $0$ indicating that the two distributions are identical, and a value greater than $0$ indicating that the two distributions are different.}. Approach \rom{2} has the lowest JSD, indicating closer $SR$ across rooms. RR has the highest overall $SR$ but it has the highest JSD indicating no fairness in the $SR$ among $3$ rooms. 

% added -- This should be normalized so let's not comment on it.
%Approach \rom{2} and \rom{4} have similar results due to the Budget term value is relative large than Fairness index so this Budget term can dominate the results. 

%explain table2 figures
In Table~\ref{tbl: FI_and_CoV}, we show $FI$ and $CoV$ calculated over the adverse effect($\mathbf{u}$) and the satisfaction ($SR$). The fairness index ($FI$) is a metric ranging from $0$ to $1$, where $1$ means absolute fair as explained in Section~\ref{sec:miny}. %Fairness index shows how close between each room's adverse effect(\textbf{u}). Covariance is a measure of the relationship between two variables. The metric evaluates how much the variables change together.

%explain FI CoV of u

Table~\ref{tbl: FI_and_CoV} first and second columns show that Approach \rom{1}, \rom{2}, \rom{3}, and \rom{4} have $FI_\mathbf{u}$ values close to $1$ and their $CoV_\mathbf{u}$ values are less than $0.04$. On the contrary, Mean and RR have $FI_\mathbf{u}$ around 0.97 and a $CoV_\mathbf{u}$ of 0.16. %indicating adverse effect(\mathbf{u}) is relatively fair compared with Mean and RR since optimization constrains \textbf{u}. 

\subsection{Comparison between these approaches}
Based on these analysis from Tables~\ref{tbl: u_and_sat},~\ref{tbl: FI_and_CoV}, and~\ref{tbl:stats} we observe that Approach \rom{2}, and Approach \rom{5} provide the best results in terms of $FI_\mathbf{u}$, and $CoV_\mathbf{u}$, while Approach~\rom{1} provides better results in terms of $FI_{\mathbf{SR}}$, and $CoV_{\mathbf{SR}}$.

\subsection{Compare with the state of the art FaiRIoT~\cite{elmalaki2021fair}}

\begin{figure}[!t]
\centering
\includegraphics[width=\columnwidth]{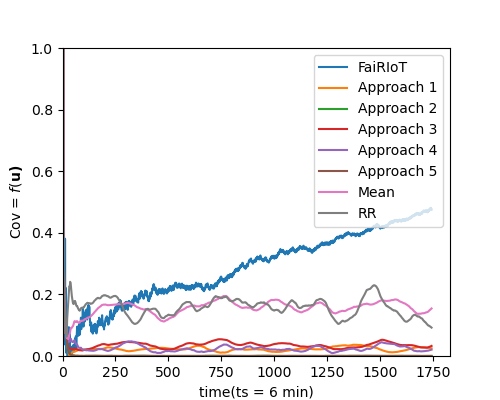}
\caption{Coefficient of Variation ($CoV$) comparison between FaiRIoT and all approaches.}\label{fig:cv_fairIoT}
\end{figure}

The closest to our approach is FaiRIoT which computes  the applied action through a weighted sum of all the desired actions by the $N$ individuals $T_a = \sum_{n=1}^N w_n T_{d_n}$. FaiRIoT uses a notion of utility %$u_{h_t} = \frac{1}{t} \sum_{j=0}^{t} \frac{j}{t} w_{h_j} $
which is the average weight assigned by a layer called ``Mediator RL'' for a particular human $h$ over a time horizon $[0:t]$. %where the factor $\frac{j}{t}$ is used to give more value to the recent weights learnt by the Mediator RL over the ones in the past.
In particular, FaiRIoT measures the fairness of the Mediator RL using the coefficient of variation ($CoV$) of the human utilities. %: $CoV = \sqrt{\frac{1}{n-1} \sum_{h=1}^{n} \frac{(u_h - \bar{u})^2}{\bar{u}^2}},$ where $\bar{u}$ is the average utility of all humans. 
The Mediator RL is said to be more fair if and only if the $CoV$ is smaller. Accordingly, in Figure~\ref{fig:cv_fairIoT}, we compare the $CoV$ in FaiRIoT with the $CoV_\mathbf{u}$ in all approaches in this paper. 
Approaches \rom{1} - \rom{5} achieve average $CoV$ around $0.20$, while FaiRIoT $CoV$ is larger than $0.6$. Approach \rom{2} and \rom{4} has the lowest $CoV$ at $0.04$. \textbf{Hence, using F$in$A approaches improves the fairness where $\mathbf{\emph{CoV}}$ is reduced by $\mathbf{66.7\%}$ on average.}

%% file: 05_conclusion.tex
\section{Discussion and Conclusion}

 Addressing fairness in decision-making not only aligns with the principles of ethical AI and responsible technology, but also highlights the importance of socially-aware CPS, as individuals are more likely to cooperate with, and ultimately accept, systems that they perceive to treat them fairly. In this paper, our approaches to formalizing F\emph{in}A within CPS decision-making capture the interplay between human preferences, the temporal dimension of adverse effects, and perceptions of fairness. Recognizing the complexities of these interactions is essential for designing more equitable Human-Cyber-Physical Systems. These approaches offer a multifaceted perspective on addressing the challenges posed by the impact of CPS control actions on diverse individuals within shared environments.